\documentclass{article}

\PassOptionsToPackage{numbers,sort&compress}{natbib}
\usepackage{subcaption}
\usepackage[table]{xcolor}
\usepackage{booktabs}
\usepackage{multirow}

\definecolor{bestblue}{HTML}{F4FCFE}

\usepackage[preprint]{neurips_2026}
\usepackage{subcaption}
\usepackage{wrapfig}
\usepackage[utf8]{inputenc}
\usepackage[T1]{fontenc}
\usepackage{hyperref}
\usepackage{adjustbox}
\usepackage{url}
\usepackage{booktabs}
\usepackage{amsfonts}
\usepackage{nicefrac}
\usepackage{microtype}
\usepackage{xcolor}
\usepackage{graphicx}
\usepackage{amsmath}
\usepackage{amssymb}
\usepackage{multirow}
\usepackage{amsthm}

\title{GenSpan: Generation-Calibrated Motion Span Priors for Multi-Verb Video Corpus Moment Retrieval}

\author{
\textbf{Yunzhuo Sun}$^{1}$ \quad
\textbf{Xinyue Liu}$^{1}$ \quad
\textbf{Yanyang Li}$^{1}$ \quad
\textbf{Nanding Wu}$^{1}$ \\
\textbf{Linlin Zong}$^{1}$ \quad
\textbf{Xianchao Zhang}$^{1}$ \quad
\textbf{Wenxin Liang}$^{1}$ \\[2mm]
$^{1}$Dalian University of Technology \\[1mm]
\texttt{sunyunzhuo@mail.dlut.edu.cn} \quad
\texttt{wxliang@dlut.edu.cn} \\[1mm]
Code \& Model: \url{https://github.com/YunzhuoSun/Manba-VMR}
}

\begin{document}

\maketitle

\begin{abstract}

Video Corpus Moment Retrieval (VCMR) aims to retrieve both the correct video and its temporal segment corresponding to a natural-language query, a task that is especially challenging for multi-verb queries where temporal action ordering is critical. Existing approaches often rely solely on text or static images and struggle to capture implicit motion dynamics, leading to retrieval errors and temporal misalignment. We propose GenSpan, a generation-calibrated VCMR framework that constructs short auxiliary videos from LLM-selected subtitle cues and decomposed sub-events, using these as temporal priors rather than direct retrieval targets. A token selector filters candidate-video features aligned with generated motion, and a bidirectional state-space model efficiently predicts video-moment tuples. Experiments on TVR and ActivityNet-Captions demonstrate that GenSpan improves corpus-level retrieval and moment localization, particularly for complex multi-action queries, while reducing computational cost compared to state-of-the-art multimodal baselines.
\end{abstract}

\section{Introduction}
\label{sec:intro}

Video Corpus Moment Retrieval (VCMR) is a practical but challenging video search task: given a natural-language query, a model must retrieve the correct video from a corpus and localize the matching temporal segment inside that video~\citep{escorcia2019temporal,lei2020tvr,liu2023vmrsurvey}. Unlike single-video moment retrieval, VCMR couples video-level ranking with moment-level boundary prediction, so a model can fail either by selecting the wrong video or by grounding the right video at the wrong time. This coupling becomes especially fragile for multi-verb queries, where the answer depends not only on what actions appear, but also on how they unfold over time.

\begin{figure}[t!]
  \centering
  \includegraphics[width=\linewidth]{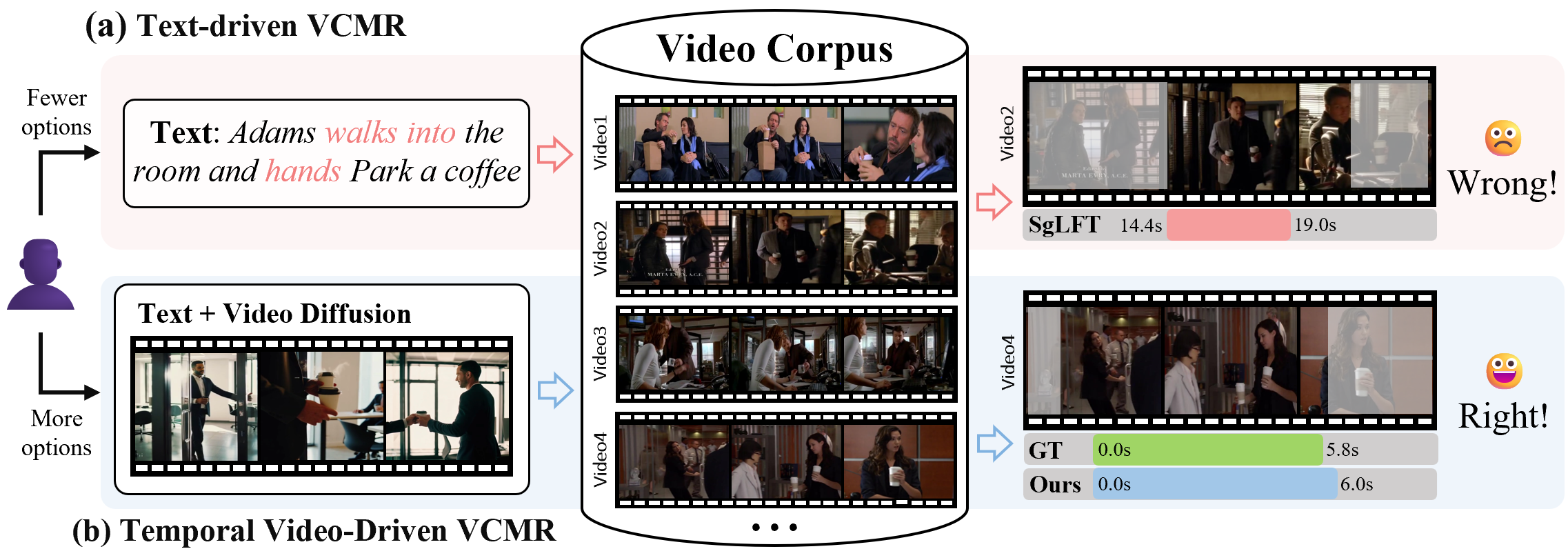}
  \caption{\small
    Motivation of temporal video-driven VCMR. (a) Text-driven VCMR relies on limited textual cues and may retrieve a wrong video-moment when a query contains ordered actions. (b) Our temporal video-driven VCMR generates an auxiliary motion prior from the query, providing richer temporal evidence for corpus-level video selection and moment localization.
  }
  \label{fig:figure1}
  \vspace{-0.3cm}
\end{figure}

Figure~\ref{fig:figure1} illustrates this difficulty. For the query ``Adams walks into the room and hands Park a coffee,'' a text-driven VCMR model must infer an ordered motion pattern from words alone: Adams first enters, then approaches, and finally hands over the coffee. Text-only matching provides limited temporal evidence and can over-emphasize isolated objects or local action words, causing the model to retrieve a visually related but temporally incorrect moment. This problem is consistent with recent observations in temporal grounding and long-video retrieval: models often struggle when queries require multi-action reasoning, implicit transitions, or fine-grained event ordering~\citep{chen2024selfsupervised,tang2025causaltsg,lan2022tsgsurvey,zhang2023tsgsurvey,yuan2025longvmr}.

One intuitive direction is to enrich the query with additional visual information. Recent multimodal-query work, such as ICQ~\citep{zhang2025multimodal}, incorporates static images including scribbles, cartoons, or realistic depictions generated by image models such as DALL-E~\citep{ramesh2022hierarchical}. These images are useful for clarifying appearance, objects, and scene semantics. However, simply adding an image does not solve the temporal-order problem: a single image can show a person holding a coffee, but it cannot express whether the person has just entered, is approaching someone, or is completing a handover. For multi-verb VCMR, the missing cue is therefore not only visual appearance, but dynamic motion structure.

We propose GenSpan, a generation-calibrated VCMR framework that uses text-to-video diffusion to construct explicit temporal priors. Instead of treating generated videos as retrieval targets, we use them as motion references. Given a query and candidate-video subtitles, an LLM first selects query-relevant subtitle cues and decomposes the query into action-centered sub-events. The fused prompt is then passed to a text-to-video model, such as CogVideoX~\citep{yang2024cogvideox}, to synthesize a short auxiliary clip that makes the implicit action order more concrete. As shown in Figure~\ref{fig:figure1}(b), this temporal prior provides more retrieval options than text alone and helps distinguish the correct video-moment from visually similar distractors.

Generated priors are informative but imperfect: they may contain background mismatch, identity mismatch, or hallucinated viewpoints. Directly concatenating all generated tokens can therefore introduce noise. To address this, GenSpan uses a generated-prior-guided token selector that keeps a measured fraction of candidate-video tokens most aligned with the auxiliary motion prior. The selected sequence is then modeled by an efficient bidirectional SSM backbone inspired by Mamba~\citep{gu2023mamba,dao2024ssm,li2024videomamba}, enabling long-sequence VCMR without the quadratic cost of dense Transformer interaction.

Our main contributions are summarized as follows:
\begin{list}{$\bullet$}{%
    \setlength{\leftmargin}{1.2em}
    \setlength{\labelwidth}{0.8em}
    \setlength{\labelsep}{0.4em}
    \setlength{\itemsep}{2pt}
    \setlength{\topsep}{2pt}
    \setlength{\parsep}{0pt}
    \setlength{\partopsep}{0pt}}
    \item A subtitle-enhanced text-to-video generation strategy that constructs explicit temporal priors for multi-verb VCMR, making implicit action order more concrete than text-only or static-image query augmentation.
    \item A GenSpan token selector that filters generation-induced noise and preserves motion-relevant candidate-video evidence before efficient bidirectional SSM modeling.
    \item Strong empirical performance on TVR, with consistent improvements over competitive VCMR and multimodal-query baselines, especially on multi-verb queries.
\end{list}

\section{Related Work}
\label{sec:related_work}

\subsection{Video Corpus Moment Retrieval}
Video moment retrieval localizes query-relevant segments in untrimmed videos~\citep{hendricks2017localizing,hendricks2018localizing,krishna2017dense}, while VCMR additionally retrieves the target video from a corpus~\citep{escorcia2019temporal,lei2020tvr}. TVR introduced video-subtitle moment retrieval and the XML baseline~\citep{lei2020tvr}. Later methods improve query-aware ranking and debiasing, including CONQUER~\citep{hou2021conquer}, SQuiDNet~\citep{yoon2022squidnet}, CTDL~\citep{yoon2023ctdl}, CKCN~\citep{chen2024ckcn}, PREM~\citep{hou2024improvingvideocorpusmoment}, EventFormer~\citep{hou2024eventformer}, and SgLFT~\citep{chen2024sglft}. These methods strengthen observed text/video/subtitle alignment; GenSpan instead injects generated motion priors for ordered multi-action queries.

\subsection{Multimodal and Generative Query Priors}
Multimodal-query work such as ICQ~\citep{zhang2025multimodal} shows that reference images can clarify visual concepts, but static images encode appearance rather than temporal transitions. Text-to-video models, including Video Diffusion Models~\citep{ho2022video}, Make-A-Video~\citep{singer2023make}, Stable Video Diffusion~\citep{blattmann2023stablevideo}, VideoCrafter~\citep{chen2023videocrafter}, and CogVideoX~\citep{yang2024cogvideox}, provide a way to synthesize action-level motion cues. We use generated clips only as auxiliary retrieval priors and filter them because generation can introduce appearance mismatch or hallucinated viewpoints.

\subsection{Efficient Long-Sequence Video Modeling}
VCMR scores many candidate videos, and generation-augmented inputs further increase sequence length. Mamba~\citep{gu2023mamba} and structured state-space duality~\citep{dao2024ssm} enable linear-time sequence modeling, with visual variants such as VideoMamba~\citep{li2024videomamba} and Vision Mamba~\citep{zhu2024vision}. GenSpan uses a bidirectional SSM after token selection: the key control variable is the generated-token keep ratio rather than dense fusion.

\begin{figure*}[t!]
  \centering
  \includegraphics[width=0.9\linewidth]{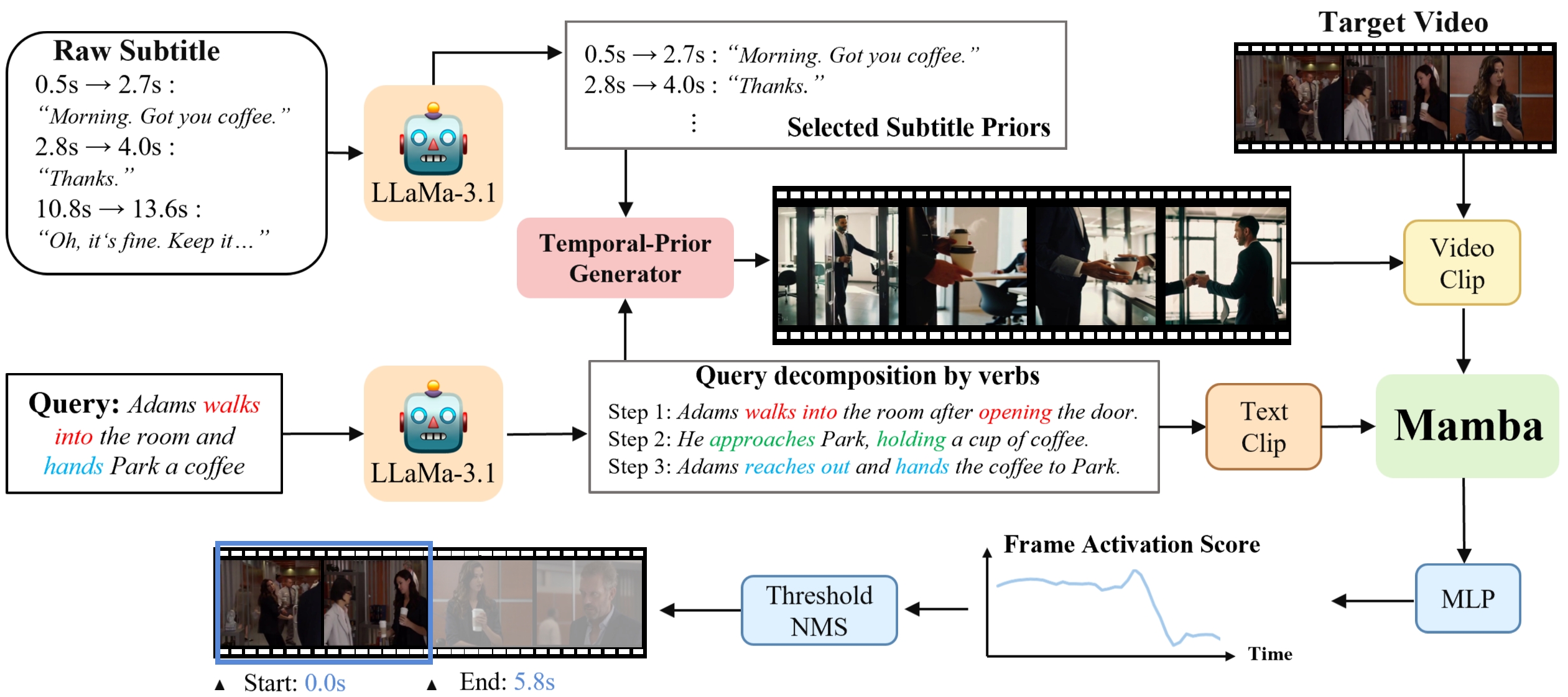}
  \caption{\small
    Overview of GenSpan. LLaMA-3.1 selects subtitle cues and decomposes the query; a temporal-prior generator synthesizes a short auxiliary video; GenSpan selects generation-aligned candidate-video tokens and predicts VCMR tuples with a bidirectional SSM.
  }
  \label{fig:method}
  \vspace{-0.25cm}
\end{figure*}

\section{Method}

In this section, we first define the video corpus moment retrieval task and provide an overview of our proposed framework in Sec.~\ref{subsec:overview}. We then detail its key components in Sec.~\ref{subsec:query_debiasing}--\ref{subsec:train}.

\subsection{Overview}
\label{subsec:overview}

Given a video corpus $\mathcal{C}=\{(V_o^n,S^n)\}_{n=1}^{N_v}$, where $V_o^n=\{v_i^n\}_{i=1}^{L_n}$ is the $n$-th untrimmed video and $S^n=\{s_j^n\}_{j=1}^{N_s^n}$ denotes its subtitles, Video Corpus Moment Retrieval (VCMR) aims to retrieve both the relevant video and the temporal moment for a textual query $Q=\{q_k\}_{k=1}^{L_q}$. The output is a ranked list of video-moment tuples
\[
\mathcal{Y}=\{(n,t^s,t^e)\},
\]
where $n$ is the video index and $t^s,t^e$ denote the start and end timestamps. VMR is a special case where the target video is given, while Video Retrieval (VR) only ranks the video index $n$ without requiring temporal boundaries.

Fig.~\ref{fig:method} illustrates our proposed framework. For each candidate video, we first apply LLaMA-3.1~\citep{grattafiori2024llama3} to match query-relevant subtitles from $S^n$, generating a refined set $S^{\prime n}\subseteq S^n$. These subtitles are fused with $Q$ to produce a composite prompt for text-to-video diffusion (e.g., CogVideoX~\citep{yang2024cogvideox}), yielding a short auxiliary video $V_g^n \in \mathbb{R}^{L_g \times d}$ as a motion prior, where $L_g \ll L_n$. The augmented input, comprising $Q$, $V_g^n$, and $V_o^n$, is then embedded and processed by our GenSpan module. The module uses generated-video priors to select motion-relevant candidate-video tokens and models the selected sequence with a bidirectional SSM. Finally, start/end logits and clip-wise relevance scores are combined to rank video-moment tuples for VCMR; max pooling over moment scores gives the VR score.

\subsection{LLM-Guided Subtitle Matching and Query Processing}
\label{subsec:query_debiasing}

\begin{wrapfigure}{r}{0.48\textwidth}
    \vspace{-0.4cm}
    \centering
    \includegraphics[width=0.46\textwidth]{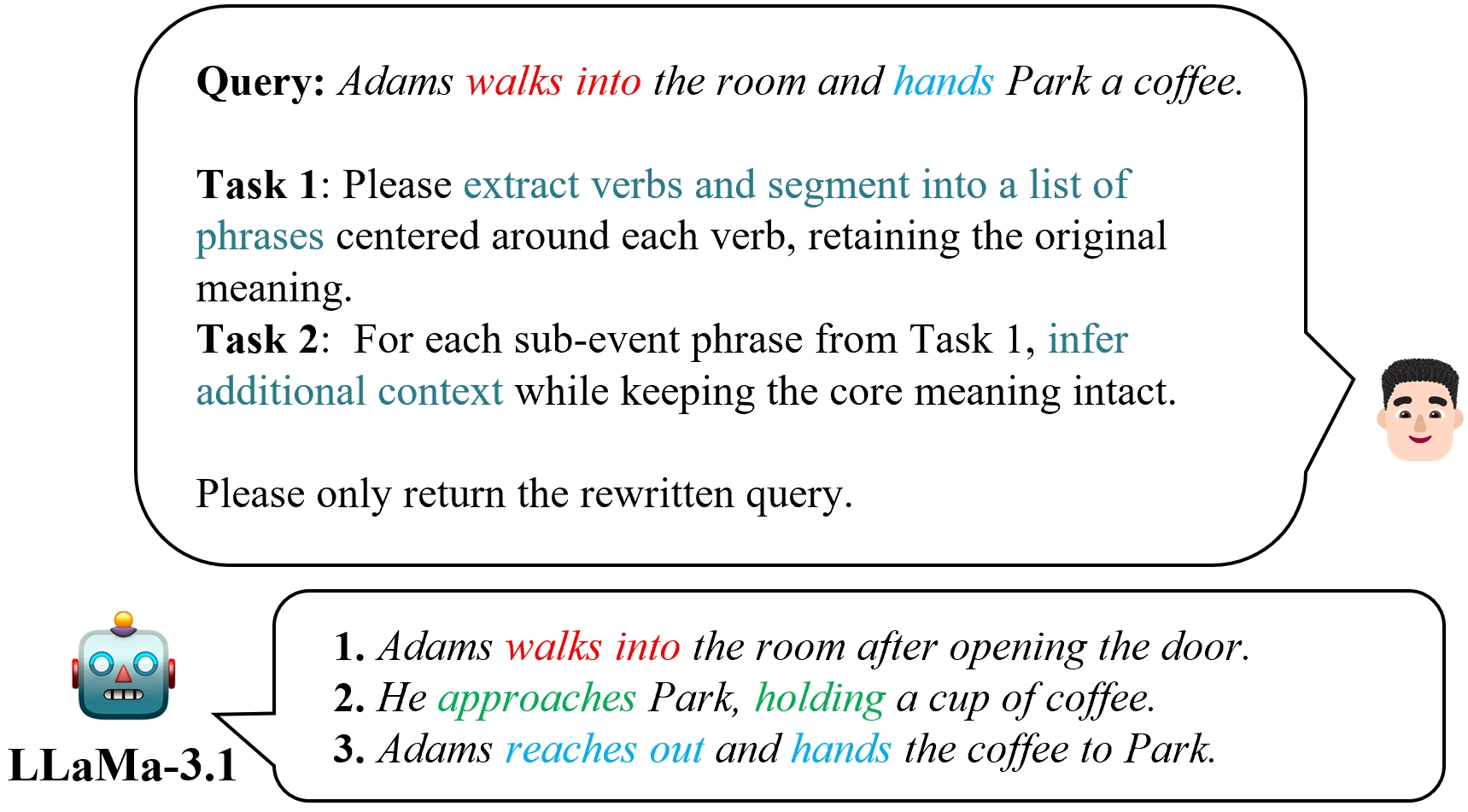}
    \caption{\small
    Example of LLM-guided query decomposition. The original query is segmented into verb-centered sub-events with inferred additional context to enrich temporal details.
    }
    \label{fig:debias}
    \vspace{-0.5cm}
\end{wrapfigure}

In video corpus moment retrieval, natural language queries (NLQs) often provide a high-level, ambiguous description of the target event, lacking fine-grained details for precise temporal grounding across many untrimmed videos. Subtitles, containing dialogue and contextual cues, offer complementary, granular information to enhance query representations. To leverage this, we employ a large language model (LLM) to process the query and match relevant subtitles for each candidate video, generating structured priors for downstream video augmentation. Rather than directly concatenating the query and subtitles, we use a structured prompt template that extracts characters, timestamped subtitle evidence, fine-grained action steps, and likely visual cues; the full template is provided in Appendix.

We adopt LLaMA-3.1~\citep{grattafiori2024llama3}, a state-of-the-art open-source LLM released in July 2024, known for its efficiency in instruction-following and text processing. The process begins by decomposing the query into action-oriented components. Specifically, we prompt the LLM to extract verbs as semantic anchors and segment the query $q$ into sub-events, inferring intermediate actions to enrich temporal sequencing. For instance, given $q = \text{"walks into the room and hands a coffee"}$, the LLM identifies verbs like \textit{walks} and \textit{hands}, producing sub-queries $q_1 = \text{"walks into the room after opening the door"}$, $q_2 = \text{"approaches Park, holding a cup of coffee"}$, and $q_3 = \text{"reaches out and hands the coffee"}$ As illustrated in Figure~\ref{fig:debias}, this decomposition supplements implicit steps (e.g., "opening the door" and "approaching"), providing richer temporal context while preserving the original meaning.

Formally, let the tokenized query be $q = [w_1, w_2, \dots, w_m]$. The LLM extracts a verb set $V = \{v_1, v_2, \dots, v_k\}$ and segments $q$ into $k$ phrases centered around each $v_j$. This is guided by the prompt: \textit{"Decompose the query '[q]' into sub-events by verbs, inferring intermediate actions while keeping the core meaning intact. Output as a list of phrases."}

For each subtitle sentence $s_j^n$ in the subtitle set 
$S^n = \{s_1^n, s_2^n, \dots, s_{N_s^n}^n\}$ of a candidate video, 
we evaluate its relevance to each sub-query $q_i$. Subtitles are processed sentence-by-sentence to maintain contextual integrity. The LLM calculates a relevance score $r_j^n$ for each subtitle $s_j^n$ against each $q_i$ using the prompt: 
\textit{"Assess if subtitle '[s\_j]' relates to query sub-event '[q\_i]'. Output a score from 0 (irrelevant) to 1 (highly relevant) and a brief reason."}

The aggregated relevance score for subtitle $s_j^n$ is computed as:
\[
r_j^n = \max_i \sigma(\text{LLM}(q_i, s_j^n)),
\]
where $\sigma(\cdot)$ normalizes the LLM output to the range $[0,1]$. We then select query-relevant subtitles with $r_j^n > \eta$, forming a refined subtitle subset $S^{\prime n} \subseteq S^n$. This matching process bridges the abstract overview in the query with fine-grained linguistic details from the subtitles (e.g., speaker-specific dialogues), and is used to guide video generation in a temporally grounded manner.

\subsection{Temporal Prior Generation via Video Diffusion}
\label{subsec:generate_candidates}

To capture hidden temporal dynamics absent in traditional text queries or static image augmentations, we generate auxiliary short videos as temporal priors. These videos are synthesized from the query fused with matched subtitles of each candidate video, providing motion-rich enhancements that bridge coarse query descriptions with fine-grained dialogue cues in subtitles.

We employ CogVideoX~\citep{yang2024cogvideox}, a state-of-the-art open-source text-to-video diffusion model released by Tencent in 2024. CogVideoX is capable of generating 6-second videos at 720$\times$480 resolution and 8 frames per second, and supports flexible text prompts via an accessible GitHub interface—making it readily integrable into our pipeline.

The generation process begins by constructing a composite prompt $p^n$ that integrates the query $q$ with the refined subtitle set $S^{\prime n}$ (as described in Sec.~\ref{subsec:query_debiasing}). Subtitles provide granular details (e.g., character dialogues) to refine the query's ambiguous overview. We prompt an LLM (LLaMA-3.1) to perform fusion via: \textit{"Combine query '[q]' with subtitles [S'] into a coherent narrative for video generation, emphasizing motion sequences."}

Formally, we define the prompt as:
\[
p^n = q \oplus \text{LLM}(\{s\}_{s \in S^{\prime n}}),
\]
where $\oplus$ denotes concatenation with transitional phrases (e.g., \textit{"as described in dialogue:"}) to ensure narrative flow, and $\text{LLM}(\cdot)$ represents the fusion of subtitle sentences into a coherent sequence. The diffusion model then samples a video:
\[
v_g^n \sim \mathcal{D}(p^n; \Theta_D),
\]
where $\mathcal{D}$ is CogVideoX parameterized by $\Theta_D$, producing a short clip that captures implicit motions such as ``walking and handing'' as a dynamic sequence.

\begin{wrapfigure}{r}{0.52\linewidth}
  \vspace{-0.4cm}
  \centering
  \includegraphics[width=\linewidth]{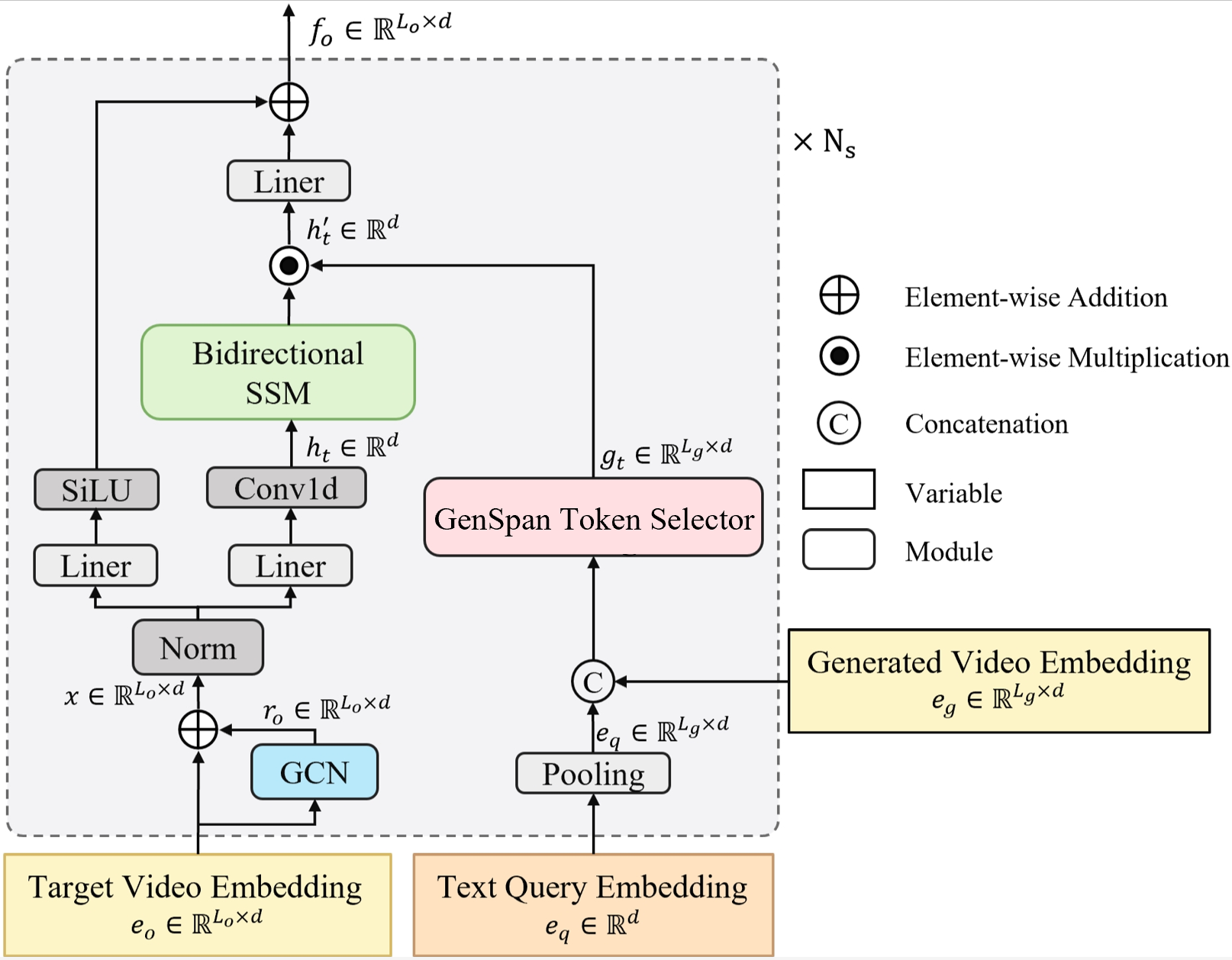}
  \caption{\small
    Architecture of the proposed GenSpan temporal module. The GenSpan Token Selector filters candidate-video tokens using text and generated-video priors, and the selected tokens are processed by a bidirectional SSM to produce contextualized features for moment prediction.
  }
  \label{fig:mamba_architecture}
  \vspace{-0.5cm}
\end{wrapfigure}

\subsection{GenSpan Token Selector and Bidirectional SSM}
\label{subsec:mamba}

Generated videos provide useful motion priors, but may also introduce background mismatch and hallucinated content. Therefore, instead of directly injecting all generated tokens into the temporal backbone, we use the generated video as a \textit{selection prior} and design a \textit{GenSpan Token Selector} to filter candidate-video tokens before sequence modeling. This reduces redundant temporal tokens, preserves motion-relevant evidence, and retains the efficiency advantage of State Space Models (SSMs) for long video sequences.

For each candidate video, the augmented input consists of the text query $q \in \mathbb{R}^{d}$, generated video $v_g^n \in \mathbb{R}^{L_g \times d}$, and original video $v_o^n \in \mathbb{R}^{L_n \times d}$, where $L_g \ll L_n$. The query is encoded by a CLIP text encoder as $e_q \in \mathbb{R}^{d}$, while the generated and original videos are encoded by a CLIP video encoder as $e_g^n \in \mathbb{R}^{L_g \times d}$ and $e_o^n \in \mathbb{R}^{L_n \times d}$. To model structural relations among candidate-video tokens, we further obtain graph-based relational embeddings $r_o^n \in \mathbb{R}^{L_n \times d}$ using a GCN on normalized frame features:
\[
x^n = e_o^n + r_o^n \in \mathbb{R}^{L_n \times d}.
\]

The GenSpan Token Selector assigns each candidate token an importance score by jointly considering text-query relevance, generated-video alignment, and temporal motion saliency. For the $t$-th token, the score is computed as
\[
s_t^n = \phi \big( x_t^n,\; e_q,\; e_g^n \big),
\]
where $\phi(\cdot)$ aggregates the three cues into a scalar score. We keep the top $\rho L_n$ tokens, where $\rho \in (0,1]$ is the keep ratio, and form
\[
x^{s,n} \in \mathbb{R}^{L_s \times d}, \qquad L_s = \rho L_n.
\]
This suppresses static background and generation-induced noise while preserving tokens related to action order and query semantics.

The selected sequence $x^{s,n}$ is then fed into a bidirectional SSM, which captures both past and future temporal contexts for boundary localization:
\[
h_t = A h_{t-1} + B x_t^{s,n}, \qquad y_t = C h_t,
\]
where $A$, $B$, and $C$ are learnable state transition matrices. Stacked bidirectional SSM layers with lightweight local operators, such as 1D convolutions and SiLU activations, produce contextualized selected features
\[
f_s^n \in \mathbb{R}^{L_s \times d}.
\]

Since the selected tokens retain their original temporal indices, we scatter $f_s^n$ back to the full $L_n$-length timeline and obtain $\tilde{f}_o^n \in \mathbb{R}^{L_n \times d}$ for prediction. A linear head outputs start logits, end logits, and clip-wise relevance scores. For candidate video $n$, the video-moment score is
\[
\psi^n(t^s,t^e)=\sigma(p_s^n(t^s))\sigma(p_e^n(t^e))
\frac{1}{t^e-t^s+1}\sum_{t=t^s}^{t^e} r_t^n,
\]
where $r_t^n$ is the clip relevance score. We rank all tuples $(n,t^s,t^e)$ by $\psi^n$ for VCMR, and obtain the VR score by $\max_{t^s,t^e}\psi^n(t^s,t^e)$.

\subsection{Loss Functions for Multi-Modal Controlled Mamba}
\label{subsec:train}
To optimize the multi-modal controlled Mamba network, we employ a combination of losses tailored to VCMR, focusing on both accurate moment boundary prediction and corpus-level relevance scoring.

Let $p_s^n, p_e^n \in \mathbb{R}^{L_n}$ denote the predicted start and end logits after scattering back to the original timeline of candidate video $n$, with ground-truth boundaries $s^\wedge, e^\wedge \in \{1, \dots, L_n\}$ for positive query-video pairs. We also compute clip-wise relevance scores $r^n \in [0,1]^{L_n}$ as:
\[
r_t^n = \sigma(W_r \tilde{f}_{o,t}^n),
\]
where $\tilde{f}_o^n$ are the contextualized features from Mamba after scattering to the original timeline, $W_r \in \mathbb{R}^{1 \times d}$ is a learnable projection, and $\sigma$ is the sigmoid function. Ground-truth relevance labels $r^{*,n} \in \{0,1\}^{L_n}$ mark clips within $[s^\wedge, e^\wedge]$ as positive for the matched video and assign all-zero labels to negative videos in the corpus. Thus, the clip-wise relevance loss also supplies supervision for VR.

The primary loss is a binary cross-entropy (BCE) for boundary classification, applied separately to start and end logits:
\[
\mathcal{L}_{\text{bound}} = \mathcal{L}_{\text{BCE}}(p_s^n, \delta_s) + \mathcal{L}_{\text{BCE}}(p_e^n, \delta_e),
\]
where $\delta_s, \delta_e \in \{0,1\}^{L_n}$ are one-hot indicators for $s^\wedge$ and $e^\wedge$, and $\mathcal{L}_{\text{BCE}}$ is the standard BCE loss.

To encourage precise relevance prediction, we add a BCE term on clip scores:
\[
\mathcal{L}_{\text{rel}} = -\frac{1}{L_n} \sum_{t=1}^{L_n} \left[ r^{*,n}_t \log r_t^n + (1 - r^{*,n}_t) \log (1 - r_t^n) \right].
\]

Finally, to regularize the fusion of generated priors, we include a contrastive loss $\mathcal{L}_{\text{cont}}$ that maximizes similarity between $e_g^n$ (generated video embeddings) and positive clips in $e_o^n$ while minimizing with negatives, using InfoNCE~\citep{vandenoord2018representation}:
\[
\mathcal{L}_{\text{cont}} = -\log \frac{\exp(\text{sim}(e_g^n, e_o^+)/\tau)}{\exp(\text{sim}(e_g^n, e_o^+)/\tau) + \sum \exp(\text{sim}(e_g^n, e_o^-)/\tau)},
\]
where $\text{sim}(\cdot,\cdot)$ is cosine similarity, $e_o^+$ are embeddings from ground-truth moments, $e_o^-$ are negatives sampled from non-overlapping clips or other videos in the corpus, and $\tau=0.07$ is the temperature.

The total loss is:
\[
\mathcal{L} = \lambda_1 \mathcal{L}_{\text{bound}} + \lambda_2 \mathcal{L}_{\text{rel}} + \lambda_3 \mathcal{L}_{\text{cont}},
\]
with weights $\lambda_1=1$, $\lambda_2=0.5$, $\lambda_3=0.1$ tuned on validation data. 

\section{Experiments}
\label{sec:experiments}

\begin{table*}[t]
\centering
\caption{\small Quantitative results on VCMR and VR. We report R@1 and R@10 under IoU thresholds 0.5 and 0.7 for VCMR, and R@1/R@10 for VR. The best results are in bold.}
\label{tab:vcmr_vr}
\scriptsize
\setlength{\tabcolsep}{2.3pt}
\renewcommand{\arraystretch}{1.05}
\resizebox{\textwidth}{!}{
\begin{tabular}{@{}l|cccc|cccc|cc|cc@{}}
\toprule
\multirow{3}{*}{\textbf{Method}}
& \multicolumn{4}{c|}{\textbf{TVR}}
& \multicolumn{4}{c|}{\textbf{ActivityNet-Captions}}
& \multicolumn{2}{c|}{\textbf{TVR}}
& \multicolumn{2}{c}{\textbf{ActivityNet-Captions}} \\
\cmidrule(lr){2-5}
\cmidrule(lr){6-9}
\cmidrule(lr){10-11}
\cmidrule(l){12-13}
& \multicolumn{4}{c|}{\textbf{VCMR}}
& \multicolumn{4}{c|}{\textbf{VCMR}}
& \multicolumn{2}{c|}{\textbf{VR}}
& \multicolumn{2}{c}{\textbf{VR}} \\
\cmidrule(lr){2-5}
\cmidrule(lr){6-9}
\cmidrule(lr){10-11}
\cmidrule(l){12-13}
& \textbf{R@1} & \textbf{R@10} & \textbf{R@1} & \textbf{R@10}
& \textbf{R@1} & \textbf{R@10} & \textbf{R@1} & \textbf{R@10}
& \textbf{R@1} & \textbf{R@10}
& \textbf{R@1} & \textbf{R@10} \\
& \textbf{IoU=0.5} & \textbf{IoU=0.5}
& \textbf{IoU=0.7} & \textbf{IoU=0.7}
& \textbf{IoU=0.5} & \textbf{IoU=0.5}
& \textbf{IoU=0.7} & \textbf{IoU=0.7}
& & & & \\
\midrule

HERO~\citep{li2020hero}
& 10.70 & 26.71 & 6.95 & 20.26
& 1.39 & 4.67 & 0.86 & 2.82
& 30.70 & 63.70 & 8.87 & 38.64 \\

XML~\citep{ge2021}
& 6.17 & 21.66 & 2.91 & 10.12
& 2.20 & 9.87 & 1.24 & 5.62
& 18.52 & 53.15 & 6.14 & 32.45 \\

ReLoCLNet
& 8.15 & 27.95 & 4.11 & 14.41
& 3.09 & 12.95 & 1.82 & 7.34
& 22.63 & 57.91 & 6.66 & 34.07 \\

SQuiDNet~\citep{yoon2022squidnet}
& 15.05 & 33.20 & 8.52 & 28.41
& 4.00 & 15.60 & 2.22 & 8.80
& 31.61 & 65.32 & 15.92 & 52.10 \\

CTDL~\citep{yoon2023ctdl}
& 15.42 & 34.10 & 8.74 & 28.72
& 4.05 & 15.94 & 2.26 & 8.96
& 31.80 & 65.90 & 16.10 & 52.68 \\

CKCN~\citep{chen2024ckcn}
& 15.20 & 33.98 & 7.92 & 22.00
& 3.88 & 14.90 & 2.18 & 8.70
& 30.95 & 64.70 & 15.88 & 52.24 \\

EventFormer~\citep{hou2024eventformer}
& 16.04 & 35.12 & 10.12 & 27.54
& 4.32 & 16.20 & 2.75 & 9.20
& 28.44 & 64.11 & 8.36 & 38.42 \\

CONQUER~\citep{hou2021conquer}
& 12.64 & 32.12 & 8.31 & 26.20
& 2.75 & 12.24 & 1.61 & 7.46
& 30.33 & 63.08 & 13.78 & 47.32 \\

PREM~\citep{hou2024improvingvideocorpusmoment}
& 14.37 & 33.59 & 8.65 & 28.86
& 3.73 & 14.40 & 1.94 & 8.49
& 26.70 & 61.87 & 14.76 & 51.30 \\

SgLFT~\citep{chen2024sglft}
& 16.82 & 35.74 & 9.24 & 29.94
& 4.25 & 16.83 & 2.32 & 9.45
& 32.08 & 67.31 & 16.58 & 55.77 \\

DMFAT
& 15.32 & 34.20 & 7.99 & 23.81
& 3.92 & 15.08 & 2.20 & 8.82
& 30.80 & 64.35 & 15.72 & 52.06 \\

ICQ~\citep{zhang2025multimodal}
& 17.02 & 36.80 & 9.80 & 30.12
& 4.45 & 16.72 & 2.48 & 9.31
& 32.39 & 67.84 & 16.85 & 55.94 \\

\midrule

VideoMamba~\citep{li2024videomamba}
& 14.96 & 33.62 & 8.58 & 27.44
& 3.86 & 15.20 & 2.12 & 8.54
& 30.90 & 64.86 & 15.64 & 51.74 \\

SpikeMba~\citep{li2024spikemba}
& 15.38 & 34.28 & 8.90 & 28.06
& 4.02 & 15.72 & 2.25 & 8.86
& 31.42 & 65.38 & 15.96 & 52.42 \\

\midrule

\rowcolor{bestblue}
\textbf{Ours}
& \textbf{18.77} & \textbf{40.56} & \textbf{12.12} & \textbf{33.31}
& \textbf{6.36} & \textbf{19.66} & \textbf{4.22} & \textbf{11.74}
& \textbf{34.48} & \textbf{71.11} & \textbf{18.73} & \textbf{57.72} \\

\bottomrule
\end{tabular}
}
\vspace{-0.2cm}
\end{table*}

\begin{table}[!b]
\centering

\begin{minipage}[t]{0.54\textwidth}
\centering
\captionof{table}{\small Quantitative results on VMR. We report R@1 and R@10 under IoU thresholds 0.5 and 0.7.}
\label{tab:vmr}
\scriptsize
\setlength{\tabcolsep}{2.2pt}
\renewcommand{\arraystretch}{1.04}
\resizebox{\linewidth}{!}{
\begin{tabular}{@{}l|cccc|cccc@{}}
\toprule
\multirow{3}{*}{\textbf{Method}}
& \multicolumn{4}{c|}{\textbf{TVR}}
& \multicolumn{4}{c}{\textbf{ActivityNet-Captions}} \\
\cmidrule(lr){2-5}
\cmidrule(l){6-9}
& \multicolumn{2}{c}{\textbf{IoU=0.5}}
& \multicolumn{2}{c|}{\textbf{IoU=0.7}}
& \multicolumn{2}{c}{\textbf{IoU=0.5}}
& \multicolumn{2}{c}{\textbf{IoU=0.7}} \\
\cmidrule(lr){2-3}
\cmidrule(lr){4-5}
\cmidrule(lr){6-7}
\cmidrule(l){8-9}
& \textbf{R@1} & \textbf{R@10}
& \textbf{R@1} & \textbf{R@10}
& \textbf{R@1} & \textbf{R@10}
& \textbf{R@1} & \textbf{R@10} \\
\midrule

HERO~\citep{li2020hero}
& 33.86 & 58.69 & 10.15 & 34.00
& 23.97 & 37.86 & 10.66 & 23.59 \\

CONQUER~\citep{hou2021conquer}
& 39.02 & 67.33 & 20.89 & 47.22
& 26.32 & 61.25 & 13.24 & 40.83 \\

PREM~\citep{hou2024improvingvideocorpusmoment}
& 43.77 & 74.50 & 24.68 & 59.32
& 30.55 & 68.27 & 17.01 & 44.20 \\

SgLFT~\citep{chen2024sglft}
& 42.51 & 72.41 & 21.03 & 54.62
& 31.28 & 70.13 & 16.68 & 43.27 \\

SQuiDNet~\citep{yoon2022squidnet}
& 41.31 & 73.10 & 24.74 & 58.90
& 30.70 & 69.20 & 17.20 & 43.85 \\

ICQ~\citep{zhang2025multimodal}
& 44.13 & 75.27 & 24.08 & 59.23
& 31.45 & 70.88 & 17.93 & 44.31 \\

VideoMamba~\citep{li2024videomamba}
& 40.56 & 70.98 & 20.29 & 53.61
& 30.18 & 68.16 & 15.96 & 42.37 \\

SpikeMba~\citep{li2024spikemba}
& 41.17 & 71.64 & 22.82 & 55.12
& 30.94 & 69.02 & 16.42 & 43.05 \\

\rowcolor{bestblue}
\textbf{Ours}
& \textbf{46.82} & \textbf{78.62} & \textbf{26.78} & \textbf{63.42}
& \textbf{33.24} & \textbf{73.63} & \textbf{19.92} & \textbf{47.33} \\

\bottomrule
\end{tabular}
}
\end{minipage}
\hfill
\begin{minipage}[t]{0.43\textwidth}
\centering
\captionof{table}{\small Results on TVR Multi-Verb queries containing $\geq 3$ verbs.}
\label{tab:multi_verb}
\scriptsize
\setlength{\tabcolsep}{3.0pt}
\renewcommand{\arraystretch}{1.04}
\resizebox{\linewidth}{!}{
\begin{tabular}{@{}l|cccc@{}}
\toprule
\multirow{2}{*}{\textbf{Method}}
& \multicolumn{2}{c}{\textbf{IoU=0.5}}
& \multicolumn{2}{c}{\textbf{IoU=0.7}} \\
\cmidrule(lr){2-3}
\cmidrule(l){4-5}
& \textbf{R@1} & \textbf{R@10}
& \textbf{R@1} & \textbf{R@10} \\
\midrule

HERO~\citep{li2020hero}
& 17.54 & 43.20 & 8.21 & 29.75 \\

CONQUER~\citep{hou2021conquer}
& 19.86 & 44.20 & 11.75 & 31.44 \\

PREM~\citep{hou2024improvingvideocorpusmoment}
& 22.74 & 49.12 & 14.08 & 36.33 \\

SgLFT~\citep{chen2024sglft}
& 20.71 & 47.86 & 12.90 & 35.40 \\

SQuiDNet~\citep{yoon2022squidnet}
& 21.81 & 49.60 & 13.06 & 35.84 \\

ICQ~\citep{zhang2025multimodal}
& 25.13 & 54.85 & 16.41 & 39.92 \\

VideoMamba~\citep{li2024videomamba}
& 21.17 & 48.52 & 12.84 & 34.91 \\

SpikeMba~\citep{li2024spikemba}
& 22.53 & 50.27 & 14.46 & 36.81 \\

\rowcolor{bestblue}
\textbf{Ours}
& \textbf{37.62} & \textbf{66.62} & \textbf{25.86} & \textbf{48.33} \\

\bottomrule
\end{tabular}
}
\end{minipage}

\vspace{-0.15cm}
\end{table}

\begin{table}[t]
\centering
\caption{\small Ablation studies on TVR. Default or best settings are highlighted in light blue.}
\label{tab:ablation_grid}
\scriptsize
\setlength{\tabcolsep}{3.5pt}
\renewcommand{\arraystretch}{1.02}

\begin{subtable}[t]{0.32\linewidth}
\centering
\caption{Main components.}
\begin{tabular}{@{}lcc@{}}
\toprule
\textbf{Method} & \textbf{R1@0.5} & \textbf{R1@0.7} \\
\midrule
\rowcolor{bestblue}
Full model & \textbf{46.82} & \textbf{26.78} \\
No LLM query proc. & 41.78 & 23.08 \\
No gen. video prior & 40.38 & 21.72 \\
No GenSpan selector & 43.04 & 23.66 \\
\bottomrule
\end{tabular}
\label{tab:main_ablation}
\end{subtable}
\hfill
\begin{subtable}[t]{0.32\linewidth}
\centering
\caption{LLM processing.}
\begin{tabular}{@{}lcc@{}}
\toprule
\textbf{Variant} & \textbf{R1@0.5} & \textbf{R1@0.7} \\
\midrule
\rowcolor{bestblue}
Full processing & \textbf{46.82} & \textbf{26.78} \\
No verb decomp. & 44.28 & 24.76 \\
No subtitle match & 43.53 & 24.21 \\
CLIP subtitle match & 44.66 & 24.92 \\
\bottomrule
\end{tabular}
\label{tab:llm_ablation}
\end{subtable}
\hfill
\begin{subtable}[t]{0.32\linewidth}
\centering
\caption{Prior source.}
\begin{tabular}{@{}lcc@{}}
\toprule
\textbf{Variant} & \textbf{R1@0.5} & \textbf{R1@0.7} \\
\midrule
\rowcolor{bestblue}
CogVideoX~\cite{yang2024cogvideox} & \textbf{46.82} & \textbf{26.78} \\
Stable Video Diff.~\cite{blattmann2023stable} & 41.75 & 24.48 \\
VideoCrafter~\cite{chen2023videocrafter} & 42.38 & 24.71 \\
DALL-E image~\cite{ramesh2022hierarchical} & 41.07 & 23.00 \\
\bottomrule
\end{tabular}
\label{tab:prior_source_ablation}
\end{subtable}

\vspace{1.2mm}

\begin{subtable}[t]{0.32\linewidth}
\centering
\caption{Generated-token keep ratio $\rho$.}
\begin{tabular}{@{}ccc@{}}
\toprule
\textbf{$\rho$} & \textbf{R1@0.5} & \textbf{R1@0.7} \\
\midrule
20\% & 44.47 & 24.51 \\
30\% & 46.34 & 26.41 \\
\rowcolor{bestblue}
\textbf{33\%} & \textbf{46.82} & \textbf{26.78} \\
40\% & 46.42 & 26.33 \\
50\% & 45.91 & 25.85 \\
\bottomrule
\end{tabular}
\label{tab:token_keep_ratio}
\end{subtable}
\hfill
\begin{subtable}[t]{0.32\linewidth}
\centering
\caption{Loss functions.}
\begin{tabular}{@{}lcc@{}}
\toprule
\textbf{Variant} & \textbf{R1@0.5} & \textbf{R1@0.7} \\
\midrule
\rowcolor{bestblue}
Full loss & \textbf{46.82} & \textbf{26.78} \\
No $\mathcal{L}_{\text{cont}}$ & 42.72 & 23.20 \\
No $\mathcal{L}_{\text{rel}}$ & 43.38 & 23.97 \\
No $\mathcal{L}_{\text{bound}}$ & 37.29 & 19.52 \\
\bottomrule
\end{tabular}
\label{tab:loss_ablation}
\end{subtable}
\hfill
\begin{subtable}[t]{0.32\linewidth}
\centering
\caption{Efficiency.}
\resizebox{\linewidth}{!}{
\begin{tabular}{@{}clccc@{}}
\toprule
\textbf{Len.} & \textbf{Model} & \textbf{R1@0.5} & \textbf{R1@0.7} & \textbf{Mem.} \\
\midrule
\multirow{2}{*}{512}
& Transformer & 44.47 & 24.66 & 16.8 \\
& Ours & \textbf{46.82} & \textbf{26.78} & \textbf{8.7} \\
\midrule
\multirow{2}{*}{700}
& Transformer & 44.12 & 24.31 & 24.0 \\
& Ours & \textbf{46.67} & \textbf{26.62} & \textbf{10.9} \\
\midrule
\multirow{2}{*}{1024}
& Transformer & OOM & OOM & -- \\
& Ours & \textbf{46.48} & \textbf{26.51} & \textbf{13.4} \\
\bottomrule
\end{tabular}
}
\label{tab:efficiency_ablation}
\end{subtable}

\vspace{-0.2cm}
\end{table}

\subsection{Datasets and Metrics}
\label{subsec:dataset}

\paragraph{Datasets.}
We evaluate on TVR~\citep{lei2020tvr} and ActivityNet-Captions~\citep{krishna2017dense}. TVR contains 21{,}793 TV-show videos and 108{,}965 queries with temporal boundaries and subtitles. We train on the official split and report validation results. ActivityNet-Captions contains about 20K web videos and 100K sentence-level temporal annotations; since it has no subtitles, priors are generated from queries only.

\paragraph{Tasks and metrics.}
We evaluate VCMR, VR, and VMR. VCMR retrieves both the video and moment from the full corpus and is measured by R@K, IoU=$\mu$; a prediction is correct if one top-$K$ video-moment tuple matches the ground-truth video with IoU $\geq\mu$. VR reports video-level R@K, while VMR assumes the video is given and evaluates localization. Unless stated otherwise, we report R@1/R@10 at IoU 0.5/0.7, plus TVR queries with at least three verbs.

\subsection{Implementation Details}
\label{subsec:implementation}

GenSpan is implemented in PyTorch 2.0 and trained on 4 NVIDIA RTX 4090 GPUs. We use CLIP ViT-B/32 with 512-dimensional embeddings. LLaMA-3.1 selects query-relevant subtitles, and CogVideoX~\citep{yang2024cogvideox} generates 6-second clips at 8 FPS offline. GenSpan uses 4 bidirectional SSM layers with hidden dimension $d=512$, state size $N=16$, and keep ratio $\rho=33\%$. We train with AdamW for 20 epochs, learning rate $1\text{e}{-4}$, and batch size 32. Loss weights are $1.0/0.5/0.1$ for boundary, relevance, and contrastive losses. For VCMR, we rank videos, localize moments in top candidates, and apply NMS.

\subsection{Comparison with the State-of-the-Arts}
\label{subsec:comparison}

\paragraph{VCMR and VR.}
Table~\ref{tab:vcmr_vr} reports corpus-level results. On TVR VCMR, GenSpan reaches 18.77 R@1/IoU=0.5 and 12.12 R@1/IoU=0.7, outperforming SgLFT and ICQ. On ActivityNet-Captions, it also achieves the best VCMR results, with 6.36 R@1/IoU=0.5 and 4.22 R@1/IoU=0.7 without subtitles. For VR, GenSpan obtains 34.48/71.11 R@1/R@10 on TVR and 18.73/57.72 on ActivityNet-Captions, showing gains in ranking and localization.

\paragraph{VMR.}
Table~\ref{tab:vmr} evaluates the single-video setting. GenSpan improves TVR R@1 from ICQ's 44.13 to 46.82 at IoU=0.5 and from SQuiDNet's 24.74 to 26.78 at IoU=0.7. On ActivityNet-Captions, GenSpan improves over ICQ by 1.79 points at R@1/IoU=0.5 and by 3.02 points at R@10/IoU=0.7. These gains confirm that the method remains useful even when the correct video is already known, because the generated video prior supplies temporal order cues that static text or image augmentation cannot provide.

\paragraph{Multi-verb queries.}
Table~\ref{tab:multi_verb} isolates TVR queries with at least three verbs. GenSpan achieves 37.62 R@1/IoU=0.5 and 25.86 R@1/IoU=0.7, substantially outperforming ICQ by 12.49 and 9.45 points, respectively. This supports our central claim: generated motion priors are most beneficial when the query describes multiple ordered actions rather than a single static event.

\subsection{Ablation Studies}\label{subsec}Table~\ref{tab:ablation_grid} summarizes six ablation groups on TVR.

\paragraph{Main components.}Removing LLM-based query and subtitle processing drops R@1/IoU=0.5 from 46.82 to 41.78, showing that decomposing multi-verb queries and aligning subtitles are important for corpus-level retrieval. Removing the generated video prior causes a larger drop to 40.38, confirming that dynamic priors provide information beyond text and subtitles. Removing the GenSpan selector is also harmful: direct concatenation retains more generated noise and reduces R@1/IoU=0.5 to 43.04.

\paragraph{LLM processing.}The LLM-processing ablation shows that verb decomposition and subtitle matching are complementary: removing either one hurts both loose and strict IoU. Replacing LLaMA-based matching with CLIP matching is better than removing subtitles entirely but still below the full LLM-guided pipeline, suggesting that instruction-based reasoning is useful for aligning subtitles with multi-step queries.

\paragraph{Prior source and token selection.}CogVideoX produces the strongest temporal priors among the tested generators. Stable Video Diffusion and VideoCrafter remain helpful but trail CogVideoX, while DALL-E image priors are clearly weaker because they cannot express motion order. The generated-token keep-ratio ablation shows that retaining 33

\paragraph{Losses and efficiency.}
All three losses contribute. Removing contrastive, relevance, or boundary loss weakens alignment, clip discrimination, or localization, with boundary loss most important. Replacing bidirectional SSMs with a Transformer increases memory and causes OOM at length 1024, whereas GenSpan remains stable with 13.4GB.

\subsection{Further Analysis}
\label{subsec:analysis}

\begin{figure}[t]
  \centering
  \vspace{-0.2cm}
  \includegraphics[width=\linewidth]{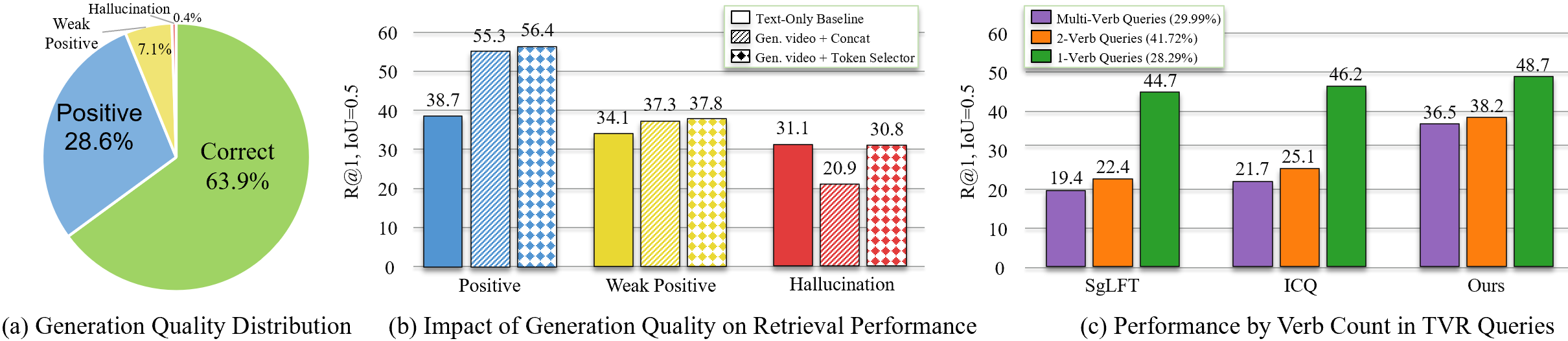}
  \caption{\small Generation quality and verb-count analysis on TVR.}
  \label{fig:gen_quality_analysis}
  \vspace{-0.2cm}
\end{figure}

\paragraph{Generation quality.}
We annotate generated priors as Correct, Positive, Weak Positive, and Hallucination. Correct clips match the query; Positive clips preserve actions but mismatch scenes; Weak Positive clips contain partial action errors; Hallucination denotes missing semantics from poor viewpoints or partial bodies. Fig.~\ref{fig:gen_quality_analysis}(a) shows 63.9\% Correct, 28.6\% Positive, 7.1\% Weak Positive, and 0.4\% Hallucination.

Fig.~\ref{fig:gen_quality_analysis}(b) explains why generated priors help. Positive samples provide strong guidance, improving R@1/IoU=0.5 from 38.7 for the text-only baseline to 55.3 with generated-video concatenation and 56.4 with the GenSpan selector. Weak-positive samples are less accurate but still useful, improving from 34.1 to 37.8 with the selector. We attribute this to the preservation of action temporal order, which is more important for VMR than exact background appearance. Hallucinated samples are harmful under naive concatenation, dropping performance from 31.1 to 20.9, but the GenSpan selector recovers performance to 30.8 by filtering noisy generated tokens.

\paragraph{Verb-count analysis.}
Fig.~\ref{fig:gen_quality_analysis}(c) shows that GenSpan's advantage grows with query complexity. For 1-verb queries, Ours reaches 48.7 R@1/IoU=0.5. For 2-verb and multi-verb queries, GenSpan reaches 38.2 and 36.5, substantially above SgLFT and ICQ.

\begin{figure}[t]
  \centering
  \vspace{-0.2cm}
  \includegraphics[width=\linewidth]{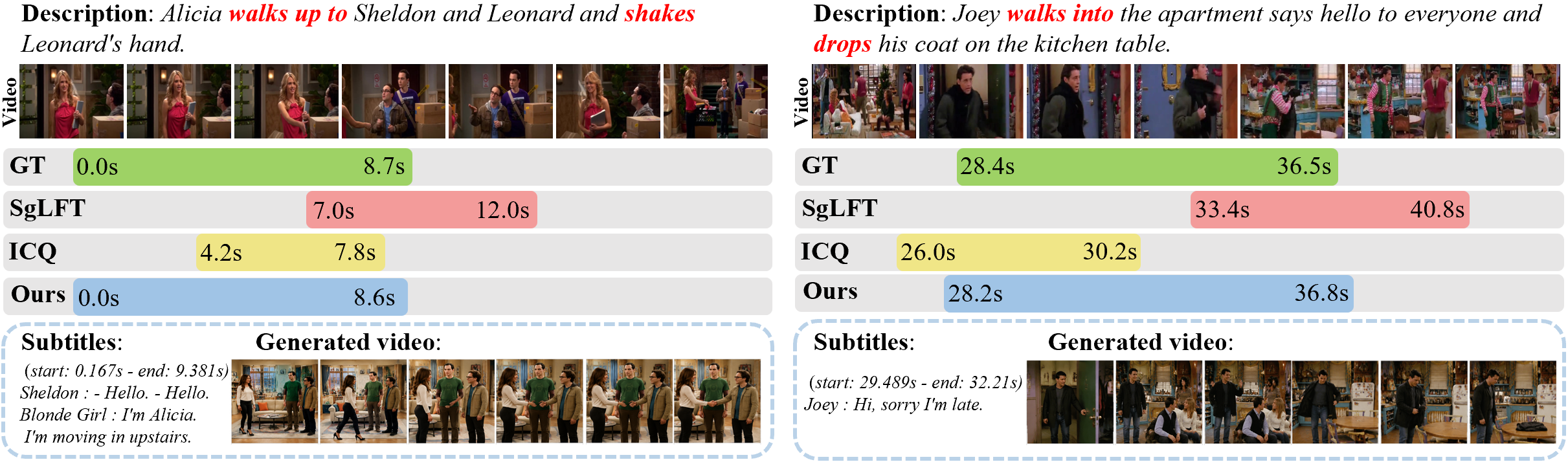}
  \caption{\small Qualitative examples on TVR.}
  \label{fig:qualitative_cases}
  \vspace{-0.2cm}
\end{figure}

\paragraph{Qualitative results.}
Fig.~\ref{fig:qualitative_cases} shows two multi-action examples: Alicia walks up to Sheldon and Leonard before shaking Leonard's hand, and Joey enters the apartment before dropping his coat. SgLFT shifts to visually salient late clips, and ICQ misses the full chain. GenSpan preserves event order and predicts intervals closer to the ground truth.
\section{Conclusion}

In this paper, we address the limitations of text-driven Video Corpus Moment Retrieval (VCMR) by proposing GenSpan, a two-stage framework that generates short temporal video priors from queries fused with relevant subtitle cues and uses them to select motion-relevant candidate-video tokens. Our approach captures hidden temporal dynamics in multi-verb queries, effectively distinguishing correct video-moment tuples from visually similar distractors. By combining LLM-guided subtitle matching, temporal-prior generation, and a bidirectional state-space backbone, GenSpan improves corpus-level retrieval and moment localization while reducing computational overhead compared to existing multimodal baselines. Experiments on TVR and ActivityNet-Captions confirm consistent gains, especially for complex multi-action queries. 

\bibliographystyle{plainnat}
\bibliography{refs_neurips}
\newpage
\appendix
This appendix provides additional details and analyses for GenSpan. Sec.~\ref{app:prompt_template} gives the structured LLM template used for query processing and subtitle matching. Sec.~\ref{app:temporal_relevance} visualizes temporal relevance maps for different model variants. Sec.~\ref{app:qualitative} provides qualitative multi-verb examples, and Sec.~\ref{app:implementation} summarizes implementation details.

\section{Prompt Template for LLM-Guided Query Processing}
\label{app:prompt_template}

\begin{figure*}[t]
  \centering
  \includegraphics[width=0.92\textwidth]{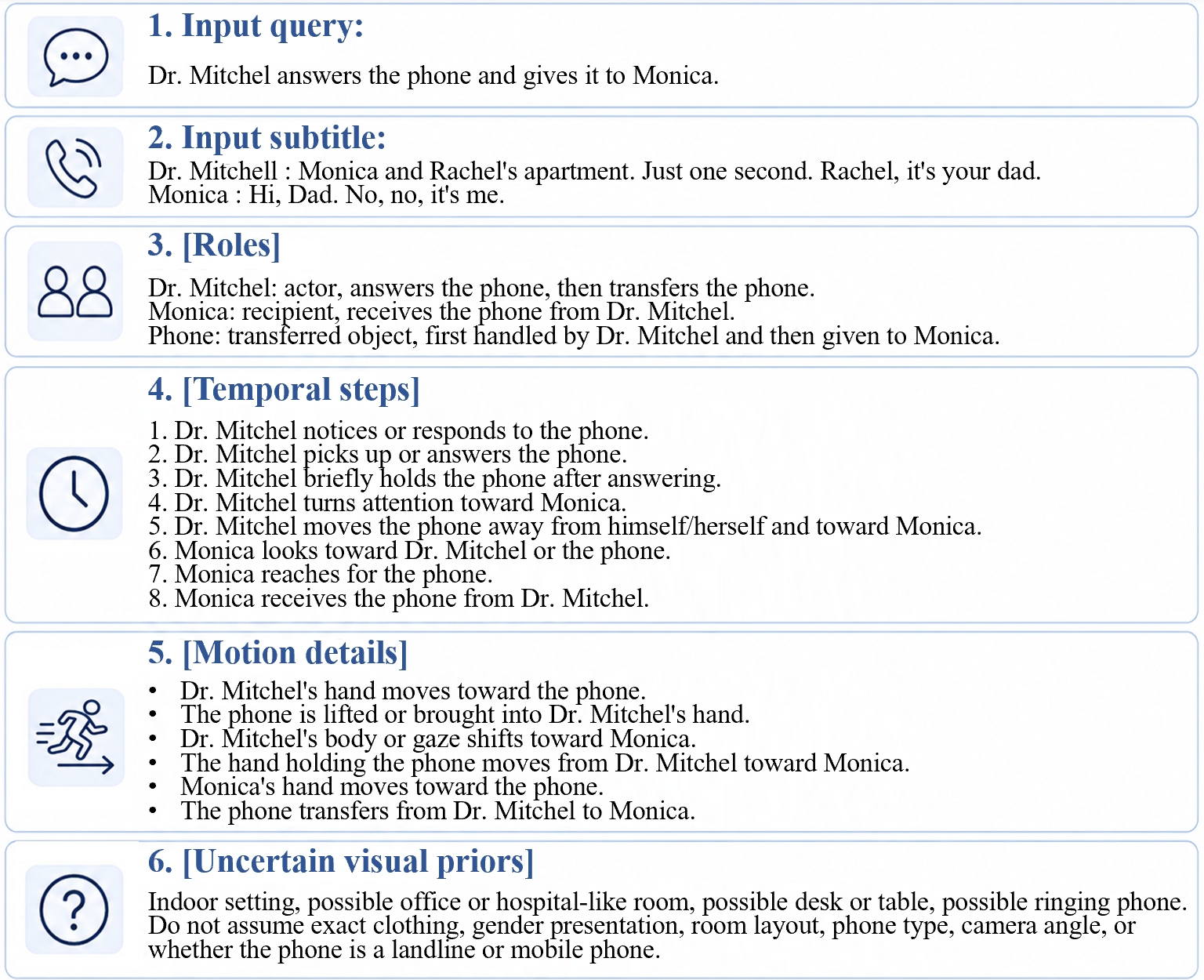}
  \caption{\small
    LLM-guided query processing template. Given the input query and matched subtitles, the LLM extracts roles, ordered temporal steps, fine-grained motion details, and uncertain visual priors. This structured representation provides clearer motion instructions than raw query-subtitle concatenation.
  }
  \label{fig:llm_prompt_template}
\end{figure*}

Our LLM module is designed to do more than directly concatenate the query and subtitles. For each candidate video, we ask the LLM to produce a structured description that identifies roles, time-aware action steps, motion details, and uncertain visual priors. Fig.~\ref{fig:llm_prompt_template} illustrates the template and an example output. This structured output is then used in two places: subtitle selection and prompt construction for text-to-video generation.

The structured format makes subtitle use more controllable. Role fields reduce ambiguity when multiple people appear in a TV episode, timestamped subtitle evidence ties the query to candidate-video context, and the temporal steps provide an explicit action order for generation. The uncertain visual priors are also important: they indicate plausible scene or object cues while explicitly discouraging unsupported assumptions about clothing, room layout, camera angle, or other details that may cause hallucination.

\section{Temporal Relevance Map}
\label{app:temporal_relevance}

\begin{figure*}[t]
  \centering
  \includegraphics[width=\textwidth]{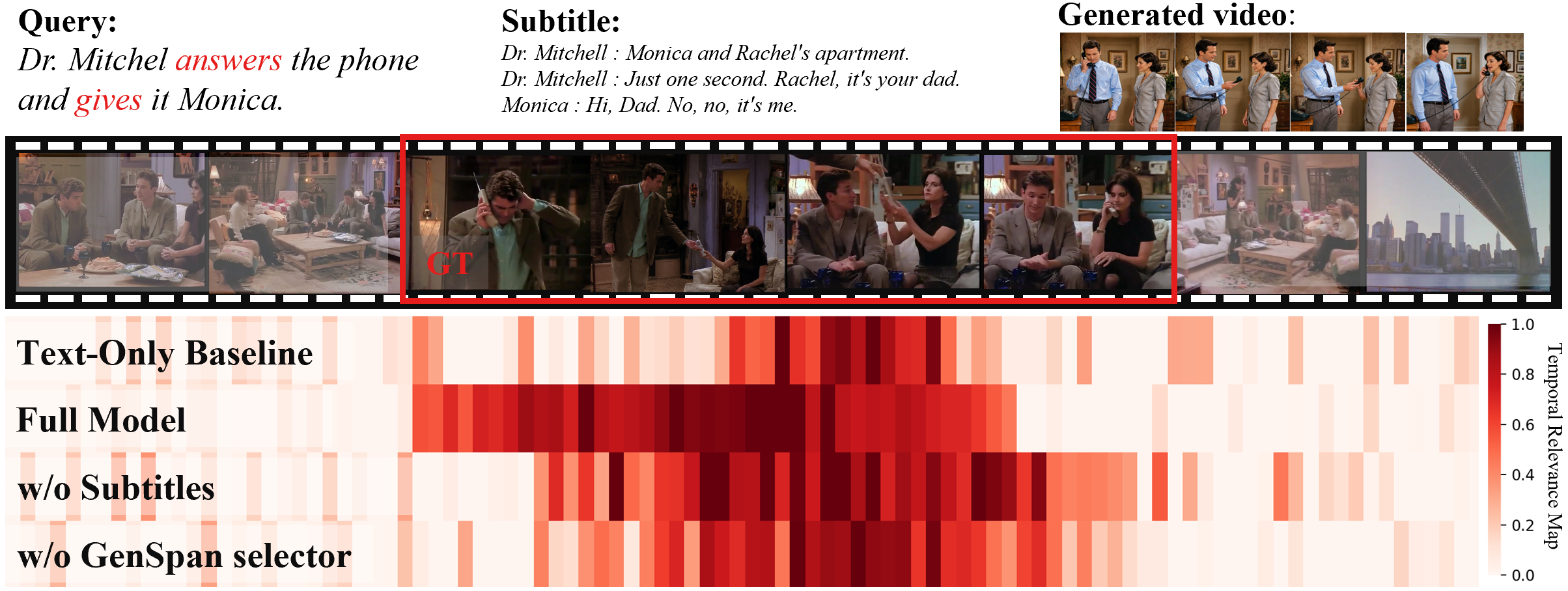}
  \caption{\small
    Temporal relevance map for different model variants. The full model concentrates high relevance around the ground-truth moment, while the text-only baseline is more diffuse. Removing subtitles weakens character- and dialogue-specific evidence, and removing the GenSpan selector produces broader high-response regions because generated-prior noise is not filtered. The heatmap denotes normalized temporal relevance intensity, not Transformer attention.
  }
  \label{fig:temporal_relevance_map}
\end{figure*}

Fig.~\ref{fig:temporal_relevance_map} visualizes temporal relevance maps for four variants: a text-only baseline, the full GenSpan model, GenSpan without subtitles, and GenSpan without the token selector. Since the temporal backbone is an SSM rather than a Transformer, these maps should not be interpreted as Transformer attention weights. For visualization, we compute a normalized temporal relevance intensity from the model's token-selection scores and clip-level relevance logits. For variants without the selector, we use the normalized clip-level relevance logits alone. All rows are min-max normalized for visual comparison.

The full model produces the most compact and temporally aligned relevance region. The text-only baseline responds to several visually plausible but incorrect regions, showing that the query alone is insufficient for precise corpus-level grounding. Without subtitles, the model still benefits from generated motion priors, but the high-relevance region shifts because character and dialogue evidence is weaker. Without the GenSpan selector, generated priors remain useful, but irrelevant or hallucinated generated-token cues are less controlled, causing the relevance map to spread beyond the target moment.

\section{Additional Qualitative Results}
\label{app:qualitative}

\begin{figure*}[t]
  \centering
  \includegraphics[width=\textwidth]{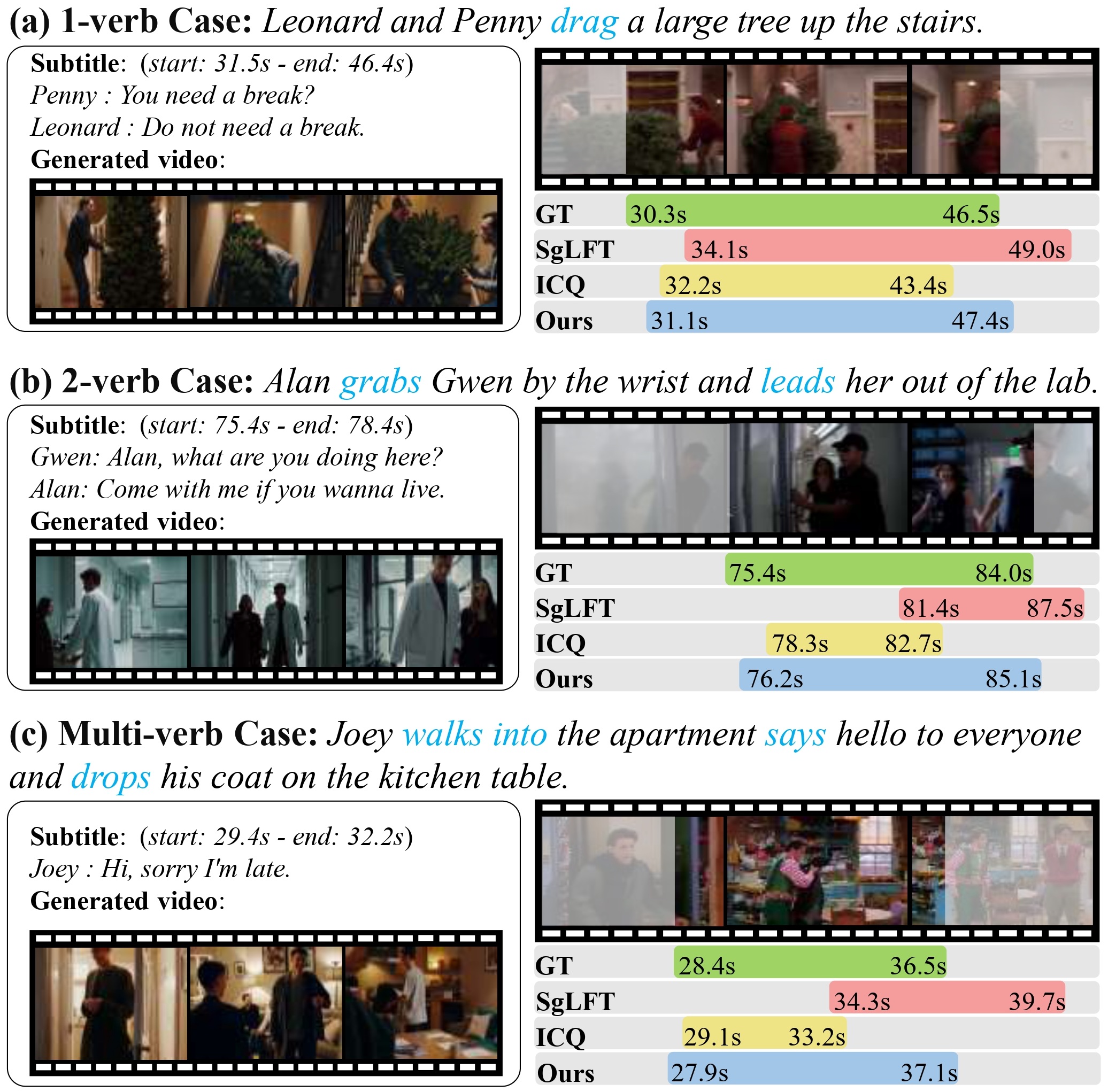}
  \vspace{-0.2cm}
  \caption{\small
    Qualitative grounding examples for 1-verb, 2-verb, and multi-verb queries. Timelines compare GT, SgLFT, ICQ, and GenSpan predictions, with subtitles and generated video frames shown for context.
  }
  \label{fig:verb_cases}
\end{figure*}

Fig.~\ref{fig:verb_cases} visualizes grounding results for queries containing one, two, or multiple verbs. The comparison shows that text- or image-augmented baselines can localize simple events, but their predictions become less stable as the query requires a longer action chain. GenSpan is more robust in the multi-verb case because the generated temporal prior explicitly represents the action order and the selector suppresses irrelevant candidate-video tokens.

\section{Implementation Details}
\label{app:implementation}

\noindent\textbf{LLM processing.}
LLaMA-3.1 is used offline for subtitle matching and query processing. For each candidate video, subtitles are provided with timestamps and speaker names when available. The LLM returns structured fields following the template in Sec.~\ref{app:prompt_template}. We then keep subtitles whose relevance score is above the threshold $\eta$ and use the structured event chain to form the generation prompt.

\noindent\textbf{Generated video prior.}
We use CogVideoX to generate 6-second auxiliary videos at 8 FPS. Generated clips are pre-computed and cached, so the retrieval model does not call the video generator during training or inference. The generated clip is used as a motion prior rather than as a retrieval target.

\noindent\textbf{Token selection.}
The GenSpan selector keeps the top $\rho L_n$ candidate-video tokens according to the generated-prior-guided relevance score. The default keep ratio is $\rho=33\%$, selected on the validation set, and it controls how many generated-prior-aligned candidate tokens are retained. Selected tokens retain their original temporal indices and are scattered back to the full timeline before start/end and clip-relevance prediction.

\noindent\textbf{Temporal relevance visualization.}
For Fig.~\ref{fig:temporal_relevance_map}, each heatmap row is min-max normalized independently. For the full model and the w/o subtitles variant, we visualize the product of normalized selector scores and clip-level relevance scores. For the text-only and w/o selector variants, we visualize normalized clip-level relevance scores. This gives a consistent qualitative view of where each model places temporal evidence.
\clearpage


\end{document}